\documentclass[letterpaper, 10 pt, conference]{ieeeconf} 

\usepackage{graphicx}
\usepackage{amsmath}
\usepackage{amssymb}

\IEEEoverridecommandlockouts                              
                                                          
\title{\LARGE \bf
Deformable State Estimation for Autonomous Surgical Tissue Retraction Under Partial Observability
}

\author{Everest Yang$^{1*}$, Skye Thompson$^{1}$, George D. Konidaris$^{1}$
\thanks{$^{1}$Brown University, Providence, RI, USA}
\thanks{$^{*}$Correspondence to everest\_yang@brown.edu}%
}

\begin{document}

\maketitle
\thispagestyle{empty}
\pagestyle{empty}


\begin{abstract}

Surgical tissue retraction requires effective manipulation planning under partial and noisy perception. We study state estimation for deformable tissue retraction, where only sparse observations of the tissue surface are available at decision time. We propose a learned state estimator that reconstructs the full deformable mesh state from 40 noisy vertex observations. The estimator combines a multilayer perceptron with a low-dimensional PCA latent representation and is trained using geometry-aware regularization that encourages smooth and physically plausible deformations. We evaluate the approach in a 2D deformable sheet simulation using single-step and multi-step retraction planning. Results show that the learned estimator achieves 98.1\% of oracle performance in multi-step retraction while supporting efficient inference. These results demonstrate that learned, geometry-regularized state estimation can support effective deformable manipulation under realistic perception constraints.

\end{abstract}

\section{INTRODUCTION}

Tissue retraction is a required step in many minimally invasive surgeries, where soft tissue must be manipulated to expose underlying anatomical targets. Autonomous robotic retraction remains challenging due to the high-dimensional, nonlinear behavior of deformable tissue and occluded perception while operating. In practice, sensing is partial and noisy, with only sparse surface observations available at decision time. Under such conditions, accurate state estimation is a prerequisite for effective planning.

Many approaches to autonomous tissue retraction assume full access to the deformable state and rely on repeated physics-based simulation to evaluate candidate actions, using finite-element or spring-based models [1, 2] or learned policies built on rich visual observations [3, 4, 5]. Surgical perception work has also explored reconstructing dense soft-tissue surfaces from stereo laparoscopy [6].

More generally, deformable state estimation and cloth modeling have been studied using approaches such as point registration with dynamic simulation [7] and Bayesian nonparametric cloth models for real-time state estimation [8]. These methods either incur substantial planning-time cost due to repeated simulation or operate in full observation regimes (e.g., dense images or stereo reconstruction) that differ from the sparse surface point observations typical of constrained surgical views.

\begin{figure}[ht]
\centering
\includegraphics[width=1.16\columnwidth]{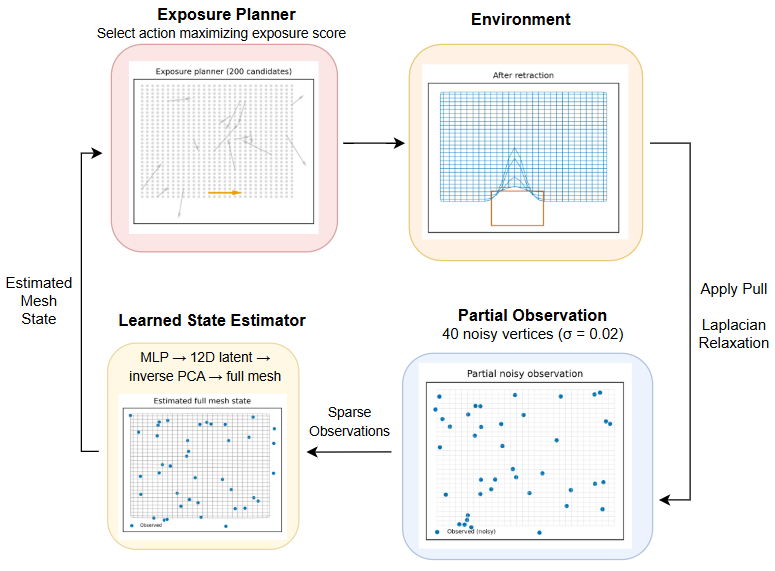}
\caption{Workflow of the proposed deformable tissue retraction system.}
\label{fig:workflow}
\vspace{-10pt}
\end{figure}

Effective planning for deformable manipulation depends critically on accurate state representation under partial and uncertain observations. Recent advances in learned dynamics and representation learning for deformable objects include diffusion-based generative state estimation for cloth manipulation [9]. Geometry-aware regularizers such as as-rigid-as-possible losses have also been shown to improve learned deformable shape generators [10]. In contrast to these approaches, we ask whether a learned state estimator operating directly on sparse noisy observations can enable effective planning for autonomous tissue retraction. The overall system pipeline is illustrated in Fig.~\ref{fig:workflow}.

\section{METHOD}

We model retracting deformable tissue to expose an underlying target region. We represent the tissue as a cloth-like sheet discretized on a regular $30 \times 30$ grid of vertices (1800-dimensional state). The sheet partially occludes a fixed target region, and the objective is to maximize target exposure through one or more retraction actions.

At each planning step, the system observes only 40 of 900 vertices (4.4\%), resampled at random each time (without replacement). Observed vertex coordinates are corrupted by Gaussian noise with standard deviation $\sigma = 0.02$, while the remaining vertices are unobserved. This partial observability makes direct planning over the deformable state infeasible.

Actions correspond to patch-based pulling operations parameterized by a grasp point on the sheet and a pull displacement. The planner selects actions to maximize an exposure score defined as the fraction of the target area not covered by the sheet. Deformation in response to actions is modeled using a diffusion-based relaxation process [11].

\subsection{Learned State Estimator}

We compute a low-dimensional latent representation of deformable tissue state using principal component analysis (PCA) [12]. PCA is applied to 5,000 randomly generated mesh states, producing a 12-dimensional latent space that captures approximately 83.7\% of the variance.

A three-layer multilayer perceptron maps the noisy, partial observations to the latent space. The network takes as input the observed vertex coordinates and their indices, and outputs a 12-dimensional latent state. The full mesh configuration is reconstructed via inverse PCA and provided to the planner.

\subsection{Geometry-Regularized Training}

The estimator is trained using a composite loss consisting of reconstruction error between the predicted and ground-truth mesh, and two geometry-aware regularization terms. Formally, the training objective is

\begin{equation}
\mathcal{L} = \mathcal{L}_{\text{recon}} 
+ \lambda_s \mathcal{L}_{\text{smooth}} 
+ \lambda_e \mathcal{L}_{\text{stretch}}.
\end{equation}

Regularization weights are fixed to $\lambda_s = \lambda_e = 0.1$ throughout. Given a reconstructed mesh $\hat{X} \in \mathbb{R}^{H \times W \times 2}$ and ground-truth mesh $X \in \mathbb{R}^{H \times W \times 2}$, we define the reconstruction loss as mean squared error over all vertices:

\begin{equation}
\mathcal{L}_{\text{recon}} 
= \frac{1}{HW} \sum_{i,j} 
\|\hat{X}_{i,j} - X_{i,j}\|_2^2.
\end{equation}

The smoothness term penalizes differences between neighboring vertices
in the reconstructed mesh:

\begin{equation}
\mathcal{L}_{\text{smooth}} 
= \frac{1}{N_{\text{nbr}}} 
\sum_{(i,j)\sim(i',j')} 
\|\hat{X}_{i,j} - \hat{X}_{i',j'}\|_2^2,
\end{equation}

where the sum is over horizontal and vertical neighbors. The stretch term penalizes deviations of edge lengths in the reconstructed mesh from those in the undeformed reference configuration. Training data is generated during training (not pre-recorded) by sampling random mesh configurations. We apply the same observation model during training, sampling 40 vertices and adding Gaussian noise (\(\sigma = 0.02\)).

\subsection{Multi-Step Planning}

We evaluate both single-step and multi-step retraction planning. In the multi-step setting, the system observes the tissue, estimates the state, selects and executes an action, then re-observes and re-estimates the state before selecting a second action. Actions are selected by sampling 200 candidate pulls and choosing the one that maximizes predicted exposure under the estimated state.

\section{EXPERIMENTS}

All experiments are conducted in simulation using a randomized 2D deformable sheet model. Deformation uses Laplacian relaxation [13]: the relaxation step size is sampled as $\alpha \sim \mathrm{Uniform}(0.15, 0.35)$, and the number of relaxation steps is sampled uniformly between 5 and 15 per episode. We evaluate performance over 200 episodes with a fixed random seed. Metrics include exposure improvement, success rate (defined as improvement greater than $0.05$), and statistical significance measured using paired and independent t-tests.


\subsection{Single-Step Results}

We compare the learned estimator against an oracle (full state access) and a naive baseline that fills observed vertex positions and leaves unobserved vertices at the undeformed reference. In single-step retraction, the learned estimator achieves a mean exposure improvement of $0.1055$ with a $50.0\%$ success rate, compared to $0.0989$ and $47.0\%$ for the oracle and $0.0951$ and $45.5\%$ for the naive baseline.

\subsection{Multi-Step Results}

Multi-step planning yields significant gains for both oracle and learned estimators. After two steps, the oracle achieves a mean improvement of $0.0958$ with a $45.0\%$ success rate, while the learned estimator achieves $0.0941$ with the same success rate, corresponding to $98.1\%$ oracle recovery. 

Relative to step-one performance, multi-step planning yields improvements of $7.3\%$ for the oracle and $15.5\%$ for the learned estimator.

\subsection{Oracle Analysis and Problem Distribution}

We perform an analysis to validate that comparison to oracle performance is an appropriate evaluation in this setting. We vary the number of sampled actions $n_\text{samples} \in \{200, 500, 1000, 2000\}$. Mean exposure improvement increases by only 2.7\% across this 10× budget range, and the success rate remains constant at 44.5\%. An independent t-test between $n_\text{samples}=200$ and $2000$ yields $p=0.835$, indicating that the oracle is effectively near-optimal even at low compute. Matching oracle performance at $n_\text{samples}=200$ is therefore a meaningful target.

To interpret these success rates, we examine the episode distribution. Approximately 57\% of episodes begin with the target already fully exposed ($\text{exposure}_0 \approx 1.0$) and are thus unimprovable under our success threshold of $\Delta > 0.05$. On the remaining 89 improvable episodes, the oracle succeeds in 100\% of cases (89/89). The reported 44.5\% success rate is computed over all 200 episodes, including unimprovable ones, confirming that the planner is near-optimal and that the primary challenge lies in state estimation under partial, noisy observations.

\section{CONCLUSION}

This paper presents a learned state estimator for autonomous surgical tissue retraction under partial and noisy perception. By combining a PCA-based latent representation with geometry-aware regularization, the estimator achieves multi-step performance statistically equivalent to a full-state oracle. These results suggest that planning-sufficient learned representations can replace expensive physical simulation for deformable manipulation under realistic perception constraints. In ongoing and future work, we plan to validate on real tissue data and investigate robustness under varying observation density and noise levels.

\addtolength{\textheight}{-12cm} 



\clearpage

\section*{ACKNOWLEDGMENT}
We thank David Paulius and members of the Brown Intelligent Robot Lab for helpful discussions and feedback.

\end{document}